\documentclass[journal,twoside,web]{ieeecolor}
\usepackage{generic}
\usepackage{cite}
\makeatletter
\let\NAT@parse\undefined
\makeatother
\usepackage{hyperref}
\hypersetup{hypertex=true,
            colorlinks=true,
            linkcolor=blue,
            anchorcolor=blue,
            citecolor=blue}
\usepackage{amsmath,amssymb,amsfonts}
\usepackage{bbding}
\usepackage{algorithmic}
\usepackage{booktabs}
\usepackage{graphicx}
\usepackage{algorithm,algorithmic}
\usepackage{textcomp}
\usepackage{xcolor}
\usepackage{multirow}
\usepackage{siunitx}
\newcommand{\ms}[2]{#1{\scalebox{0.65}{$\pm$#2}}}
\def\BibTeX{{\rm B\kern-.05em{\sc i\kern-.025em b}\kern-.08em
    T\kern-.1667em\lower.7ex\hbox{E}\kern-.125emX}}
\begin{document}
\title{KG-CMI: Knowledge graph enhanced cross-Mamba interaction for medical visual question answering}
\author{Xianyao~Zheng,
 \and
  Hong~Yu,
  \and
  Hui~Cui,
  \and
  Changming~Sun,
  \and
  Xiangyu~Li,
  \and
  Ran~Su,
  \and
  Leyi~Wei,
  \and
  Jia~Zhou,
  \and
  Junbo~Wang,
  \and
  and~Qiangguo~Jin
  \thanks{Manuscript submitted Nov. 4, 2025. This work was supported by the National Natural Science Foundation of China [Grant No. 62572401, No. 62222311, and No. 62322112], and the Key Research and Development Program of Shaanxi [Program No. 2025SF-YBXM-424]; Xianyao Zheng and Hong Yu contributed equally to this work. (Corresponding author: Qiangguo Jin)}
  \thanks{Xianyao~Zheng, Junbo~Wang and Qiangguo~Jin are with School of Software, Northwestern Polytechnical University, Shaanxi, China (e-mail: \{zhengxianyao,jbwang,qgking\}@nwpu.edu.cn).}
   \thanks{Hong Yu is with Tianjin Central Hospital of Gynecology Obstetrics, Tianjin, China (e-mail: tjszxfcrsk@163.com).}
  \thanks{Hui Cui is with Department of Computer Science and Information Technology, La Trobe University, Melbourne, Australia (e-mail: L.Cui@latrobe.edu.au).}
  \thanks{Changming Sun is with CSIRO Data61, Sydney, Australia (e-mail: changming.sun@data61.csiro.au).}
  \thanks{Xiangyu Li is with School of Computer Science and Technology, Harbin Institute of Technology, Harbin, China (e-mail: lixiangyu@hit.edu.cn).}
  \thanks{Ran Su is with School of Computer Software, Tianjin University, Tianjin, China (e-mail: ran.su@tju.edu.cn).}
  \thanks{Leyi Wei is with Centre for Artificial Intelligence driven Drug Discovery, Faculty of Applied Science, Macao Polytechnic University, Macao SAR, China (e-mail: weileyi@mpu.edu.mo).}
  \thanks{Jia Zhou is with Department of Cardiology, Tianjin Chest Hospital, Tianjin, China (e-mail: zhao\_yifei@tju.edu.cn).}
  \thanks{Qiangguo Jin is also with Yangtze River Delta Research Institute of Northwestern Polytechnical University, Taicang, China. (e-mail: qgking@nwpu.edu.cn).}
}

\maketitle

\begin{abstract}
Medical visual question answering (Med-VQA) is a crucial multimodal task in clinical decision support and telemedicine. Recent methods fail to fully leverage domain-specific medical knowledge, making it difficult to accurately associate lesion features in medical images with key diagnostic criteria.
Additionally, classification-based approaches typically rely on predefined answer sets. Treating Med-VQA as a simple classification problem limits its ability to adapt to the diversity of free-form answers and may overlook detailed semantic information in those answers. To address these challenges, we propose a knowledge graph enhanced cross-Mamba interaction (KG-CMI) framework, which consists of a fine-grained cross-modal feature alignment (FCFA) module, a knowledge graph embedding (KGE) module, a cross-modal interaction representation (CMIR) module, and a free-form answer enhanced multi-task learning (FAMT) module. The KG-CMI learns cross-modal feature representations for images and texts by effectively integrating professional medical knowledge through a graph, establishing associations between lesion features and disease knowledge. Moreover, FAMT leverages auxiliary knowledge from open-ended questions, improving the model's capability for open-ended Med-VQA. Experimental results demonstrate that KG-CMI outperforms existing state-of-the-art methods on three Med-VQA datasets, i.e., VQA-RAD, SLAKE, and OVQA. Additionally, we conduct interpretability experiments to further validate the framework's effectiveness.
\end{abstract}


\section{Introduction}
Medical visual question answering (Med-VQA), has emerged as a critical multimodal task, offering significant potential for enhancing clinical decision support and telemedicine applications. By integrating medical image analysis and natural language processing, Med-VQA enables intelligent systems to understand, interpret, and respond to medical queries based on both visual and textual information. This capability is vital in scenarios where timely and accurate medical decisions are necessary, such as in remote diagnostics, radiology, and personalized treatment planning. 

Recently, deep learning applications in Med-VQA have made significant progress~\cite{LLaVA-MedNEURIPS2023,medvlmr1pan2025}. For instance, convolutional neural networks (CNNs)~\cite{CNN} have been widely used due to their ability to extract basic visual features from medical images. However, CNNs have limitations in handling complex semantics and long-range dependencies~\cite{overcoming2019}. Leveraging the self-attention mechanism, Vision Transformers (ViTs)~\cite{dosovitskiy2021imageworth16x16words} can capture global image information more effectively, significantly enhancing the model's ability to understand complex structures and relationships in Med-VQA~\cite{miss2024}. 
These advancements have progressively enhanced the performance of Med-VQA models, enabling them to better process medical images and answer complex medical questions, driving continuous innovation in the field~\cite{li2023masked,li2023self}.

However, despite the promising applications, existing methods in Med-VQA face significant challenges. First, existing methods  struggle to effectively capture the semantic associations between subtle pathological features in medical images and linguistic information, especially in the face of complex cases involving multiple organs and diseases, where the model's reasoning ability is significantly insufficient.
Additionally, medical domain knowledge is highly specialized and complex, including extensive anatomical knowledge, disease diagnostic criteria, and treatment plans. Such knowledge is often scattered across various medical literatures and databases, and existing Med-VQA models find it difficult to fully utilize these valuable knowledge resources, leading to poor performance when handling rare cases and complex diagnostic problems. Moreover, classification-based approaches, which rely on closed-ended questions (e.g., Q: \textit{Are regions of the brain infarcted?} $\to$ A: \textit{Yes}), do not adequately address the diverse and often nuanced nature of free-form open-ended medical questions (e.g., Q: \textit{In which two ventricles can calcifications be seen on this CT scan?} $\to$ A: \textit{The 3rd ventricle and the lateral ventricles}). This limitation restricts the model's ability to adapt to the broad variety of potential responses in Med-VQA, often missing critical semantic information that could improve clinical decision-making.

To address these challenges, we propose a knowledge graph enhanced cross-Mamba interaction multi-task learning framework (KG-CMI)~\footnote{The code is publicly available at https://github.com/BioMedIA-repo/KG-CMI}
. The KG-CMI framework aims to achieve cross-modal semantic alignment and reasoning between medical images and textual questions by fully integrating medical knowledge graphs. The framework consists of a fine-grained cross-modal feature alignment (FCFA) module, a knowledge graph embedding (KGE) module, a cross-modal interaction representation (CMIR) module, and a free-form answer enhanced multi-task learning (FAMT) module. The FCFA utilizes a vision-text bidirectional contrastive learning strategy to strengthen the deep alignment of features between medical images and textual questions, enhancing cross-modal semantic consistency and accurately capturing subtle correlations between them. 
The KGE constructs knowledge graph-based medical knowledge representation, effectively integrating structured knowledge in the medical field to provide rich prior knowledge support for Med-VQA tasks. 
The CMIR proposes a cross-Mamba based interaction mechanism, which captures deep semantic correlations between images and text, improving cross-modal expressive ability. 
The FAMT proposes a multi-task learning strategy to simultaneously optimize classification tasks and knowledge reasoning tasks, improving the model's generalization ability and its ability to handle rare cases.

The main contributions of our approach are summarized as: 
\begin{itemize}
    \item To accurately associate subtle pathological features with linguistic descriptions, we propose a fine-grained cross-modal feature alignment (FCFA) module. It utilizes bidirectional contrastive learning to strengthen semantic consistency between image regions and textual tokens.
    \item To address the lack of domain-specific reasoning, we propose a knowledge graph embedding (KGE) module that dynamically retrieves structured medical knowledge. KGE module effectively bridges the semantic gap between subtle visual lesion features and abstract textual clinical concepts.
    \item Addressing the computational bottleneck of modeling long medical sequences, we introduce the cross-modal interaction representation (CMIR) module. CMIR achieves linear-complexity modeling of cross-modal dependencies, capturing deep correlations more efficiently than traditional quadratic attention mechanisms.
    \item We propose the Free-form answer enhanced multi-task (FAMT) learning module to overcome the limitations of closed-set classification. By leveraging generative auxiliary knowledge, this module significantly improves the model’s adaptability to diverse and nuanced real-world clinical queries.
    \item Extensive experiments on three benchmarks verify that each module mitigates its corresponding challenge. Grad-CAM visualizations further confirm that the model accurately focuses on question-relevant regions, which supports its reliability in clinical decision-making.
\end{itemize}

\section{Related Work}
\subsection{Medical visual question answering}
The Med-VQA task was introduced by the ImageCLEF Med-VQA 2018 competition~\cite{hasan2018overview}. It typically employs methods that jointly embed images and texts to process medical images and clinical questions. Early methods~\cite{peng2018umass,nguyen2019overcoming} relied heavily on pre-trained CNNs for image feature extraction~\cite{MMBERT}. For image features, ResNet or VGGNet were commonly used for this purpose. For text features, recurrent neural networks (RNNs)~\cite{RNN} and Transformer-based language models, such as bidirectional encoder representations from Transformers (BERT)~\cite{BERT}, were used for text feature extraction. 

Since 2021, Med-VQA has witnessed remarkable progress driven by pre-trained vision-language models and advanced fine-tuning strategies. Self-supervised learning approaches have helped address the scarcity of medical data through masked modeling and contrastive objectives. For example, M2I2~\cite{li2023self} and MUMC~\cite{li2023masked} achieved state-of-the-art results on multiple datasets by pre-training on medical image captioning tasks. Domain-specific adaptations of contrastive language-image pre-training (CLIP) architectures, such as PubMedCLIP~\cite{pubmedclip2023}, further improved visual encoders within existing pipelines, yielding gains of up to 3\% by fine-tuning on medical image-text pairs. More recently, generative approaches have emerged as a promising direction. The MMCAP~\cite{MMCAP2024} framework replaced traditional classification heads with generative layers, enhancing clinical applicability, while PeFoMed~\cite{liu2024pefomed} introduced parameter-efficient fine-tuning for multimodal large language models, outperforming GPT-4V by 26\% on closed-ended questions.

\begin{figure*}[htb]
    \centering
    \includegraphics[width=0.9\linewidth]{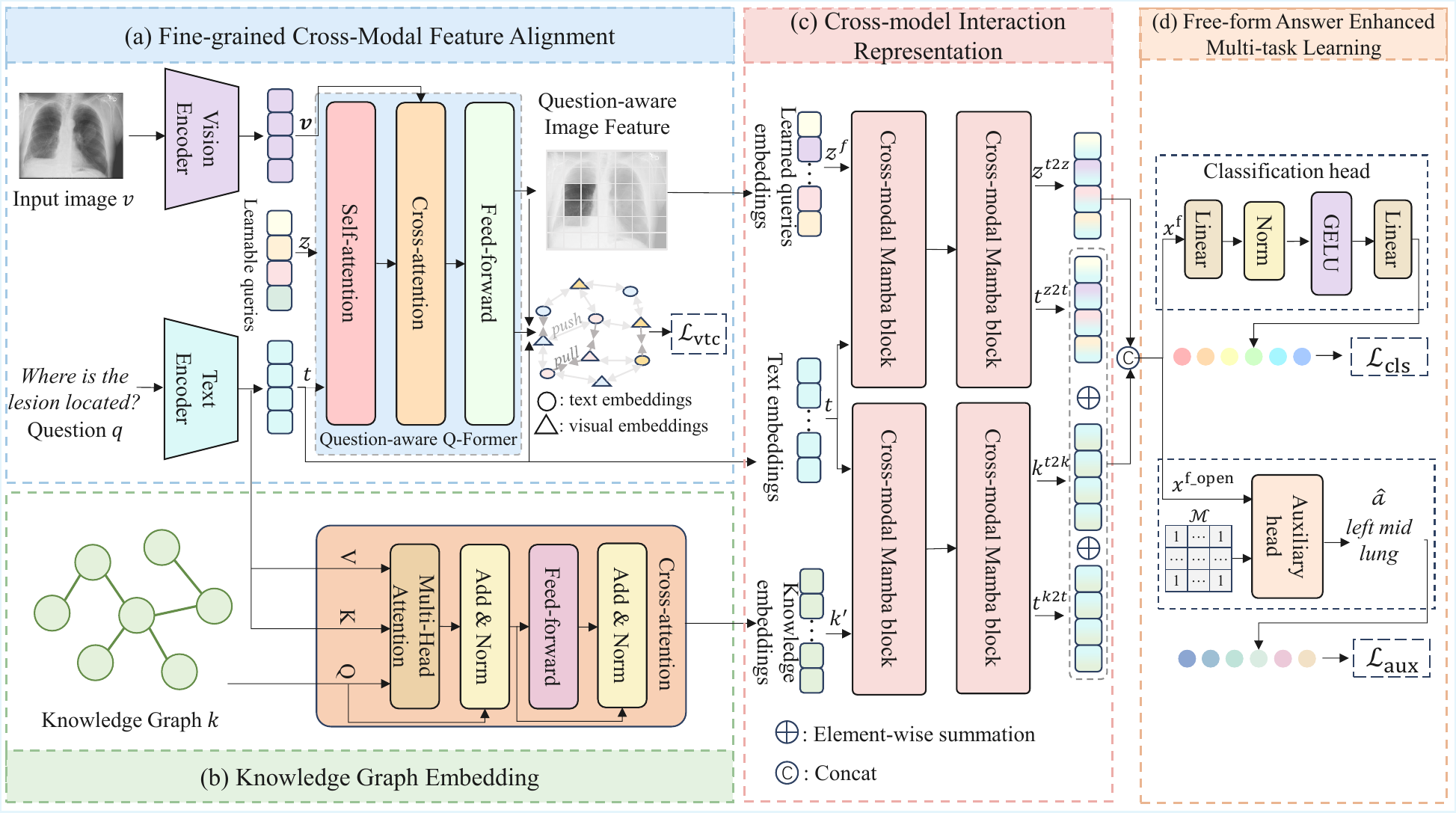}
    \caption{The overall architecture of KG-CMI. (a) Fine-grained visual-text feature alignment, (b) Knowledge graph embedding, (c) Cross-modal interaction representation, and (d) Free-form answer enhanced multi-task learning.}
    \label{fig:framework_new}
\end{figure*}

Current Med-VQA methods have achieved impressive results in answering questions related to medical images. However, these methods face limitations in effectively integrating textual medical knowledge, which is crucial for providing accurate answers. For instance, when tasked with identifying the location of a lesion, a Med-VQA model may base its response solely on the lesion's appearance in the image. In clinical practice, however, doctors consider the patient's medical history, symptoms, and other relevant factors alongside the image data to make a more accurate diagnosis.
Moreover, the image features extracted by current methods often encompass all image content, while medical images typically contain background regions and irrelevant areas. Encoding these unnecessary parts as features introduces noise and increases computational costs.
Finally, most current methods treat Med-VQA as a multi-class classification problem, relying on predefined answer sets. This approach limits the model's ability to adapt to the diversity of free-form answers and may overlook detailed semantic information.

The aforementioned limitations motivate our approach. We propose a novel framework designed to enhance the generalization ability and capacity to handle rare cases.

\subsection{Knowledge graph-based visual question answering}
The application of knowledge graphs in the medical field has become a research hotspot~\cite{yang2024lmkg,cui2025review-mkg}, as they provide structured domain knowledge that enhances both the interpretability and clinical relevance of models. Medical knowledge graphs, such as UMLS~\cite{Bodenreider2004}, contain rich information on medical entities, relationships, and attributes, thereby offering valuable support for medical research and clinical decision-making.

Significant progress has been achieved in knowledge-guided feature optimization and adaptive reasoning for Med-VQA. The grounded knowledge-enhanced medical vision-language pre-training framework~\cite{gkm2025} introduced a Transformer-based grounding module that aligns textual medical knowledge with specific anatomical regions in chest X-rays. 
More recent studies have explored dynamic knowledge retrieval and reasoning mechanisms to overcome the limitations of static knowledge integration. For example, the attentional graph neural network (AGNN)~\cite{agnn2024} employed an adaptive attention architecture that dynamically adjusts the weights of knowledge nodes based on the input question. Compared with conventional graph convolutional network (GCN) methods, AGNN reduced reasoning errors by 37\%-53\%. Building on this line of research, the KG-RAG framework~\cite{kgrag2025} integrated knowledge graph retrieval into large vision-language models (VLMs) such as LLaVA, giving rise to the KG-LLaVA model.

The field of generative Med-VQA has particularly benefited from knowledge graph integration. While the LaPA~\cite{gu2024lapa} enhanced textual question representation with knowledge graphs, PMC-VQA~\cite{pmcvqa2024} further demonstrated the effectiveness of knowledge-enhanced visual instruction tuning, achieving robust improvements in free-text answer generation across multiple medical modalities. These advancements signify a paradigm shift in research focus, moving from treating knowledge graphs as static external resources toward embedding them as dynamic reasoning components at the core of Med-VQA systems.

Our work advances this line of research by proposing a deep integration strategy that first performs structure-aware feature aggregation using graph attention network (GAT)~\cite{2018graphattentionnetworks} to encode global disease relationships. Subsequently, a cross-attention mechanism dynamically retrieves relevant knowledge from these aggregated features, guided by the specific input question.

\section{Methods}
The overall architecture of the KG-CMI framework is shown in Fig.~\ref{fig:framework_new}, primarily comprising the following core modules: fine-grained cross-modal feature alignment (FCFA), knowledge graph embedding (KGE), cross-modal interaction representation (CMIR), and free-form answer enhanced multi-task learning (FAMT).

\subsection{Problem formulation}
Similar to general VQA tasks, Med-VQA can primarily be formulated as a multi-class classification problem, applicable to both open-ended and closed-ended questions. Given a visual-text dataset $ D = \{ (v_i, q_i, a_i) \}_{i=1}^N $, where $ N $ denotes the number of training samples, each triplet $ (v_i, q_i, a_i) $ represents the medical image-question-answer triplet for the $ i $-th medical image. Additionally, we incorporate prior knowledge from an external knowledge graph $ k $ to provide the model with richer textual information. The goal of Med-VQA is to learn a mapping function $ \mathcal{F} $ to predict the optimal answer $ \hat{a} $ for a given image-question pair from the candidate answer set $ A $:  
\begin{equation}
\hat{a}_i=\arg\max_{a_i\in \mathcal{A}}\mathcal{F_{\theta}}\left(a_i|v_i,q_i,k\right),
\end{equation}
where $ \theta $ denotes the trainable parameters of the model, and $ A $ denotes the candidate answer set in $ D $.

\subsection{Fine-grained cross-modal feature alignment module}
\begin{figure}
    \centering
    \includegraphics[width=0.7\linewidth]{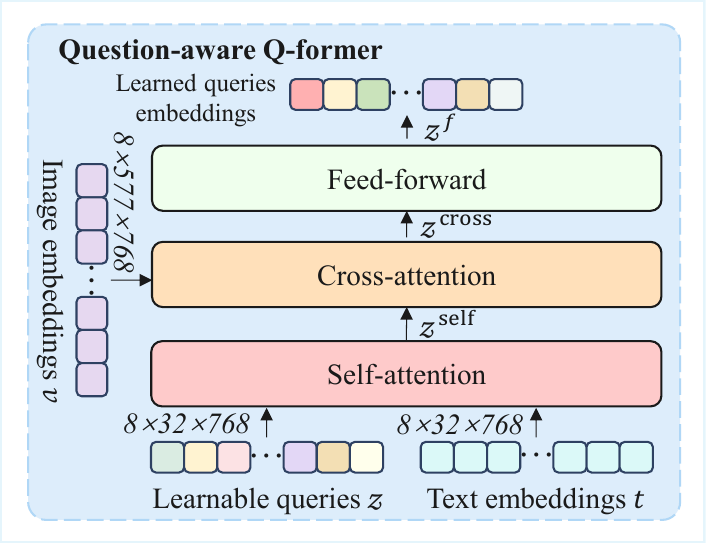}
    \caption{The detailed structure of the question-aware Q-Former (QQ-former).}
    \label{fig:module 1}
\end{figure}

As shown in Fig.~\ref{fig:framework_new}(a), we propose an FCFA module to align visual features extracted from medical images with text embeddings obtained from questions. The FCFA consists of an image encoder, a text encoder, and a question-aware Q-Former (QQ-Former). The image encoder utilizes the ViT~\cite{dosovitskiy2021imageworth16x16words} to extract visual features from medical images. The text encoder employs RoBERTa~\cite{liu2019robertarobustlyoptimizedbert} to extract semantic representations of the text questions. The QQ-Former includes a single-layer structure consisting of a self-attention layer, a cross-attention layer, and a feed-forward network, as shown in Fig.~\ref{fig:module 1}.
The QQ-Former enhances the model's ability to focus on the most relevant parts of the question in relation to the image context, thereby improving the accuracy of answering questions.

Specifically, let $\{ (v_i, q_i) \}_{i=1}^B$ denote a batch of $B$ visual-text pairs. The QQ-Former aligns the visual features $\boldsymbol{v}_i$ extracted from ViT, with the question text embeddings $\boldsymbol{t}_i$ obtained from RoBERTa. The query embeddings $\boldsymbol{z}_i$ first interact with the text features $\boldsymbol{t}_i$ through the self-attention layer, and then $\boldsymbol{v}_i$ interacts with $L$ learnable query vectors $\boldsymbol{z}_i$ via the cross-attention layer, ensuring that the query outputs incorporate detailed information throughout this process.
Let the learnable query embeddings be $\boldsymbol{z}\in\mathbb{R}^{B \times L \times d}$, where $d$ is the embedding dimension. The QQ-former is defined as follows:
\begin{equation}
    \begin{aligned}
        \boldsymbol{z}^{\text{self}} &= \text{SelfAttention}(\boldsymbol{z}),\\
        \boldsymbol{z}^{\text{cross}} &= \text{CrossAttention}(\boldsymbol{z}^{\text{self}}, \boldsymbol{v}),
    \end{aligned}
\end{equation}
where $\text{SelfAttention}$ and $\text{CrossAttention}$ are the self-attention and the cross-attention derived from the Q-former~\cite{li2023blip}. Finally, a nonlinear transformation is applied to $\boldsymbol{z}^{\text{cross}}$, typically using a multi-layer perceptron (MLP) with two hidden layers and a GELU~\cite{hendrycks2016gaussian} layer as the activation function:
\begin{equation}
   \boldsymbol{z}^{\text{f}} = \text{GELU}\left(\boldsymbol{z}^{\text{cross}} W_1 + b_1 \right) W_2 + b_2,
\end{equation}
where $W_1, W_2 \in \mathbb{R}^{d \times d}$ and $b_1, b_2 \in \mathbb{R}^d$ are learnable biases that enhance the model's nonlinear expressive ability.

To align the visual representations and text representations, we employ cross-modal contrastive learning (CMCL) to minimize the semantic discrepancy between images and texts. Let $\boldsymbol{z}^{\text{f}}\in{\mathbb{R}^{B \times L \times d}}$  denote the output query embeddings from the Q-Former, which encapsulate visual information aligned with the question. The textual features $\boldsymbol{t}_i$ are the [CLS] token embeddings (representing the global semantic information of the text) directly extracted by the RoBERTa text encoder. The contrastive loss calculates the similarity between the aligned query output $\boldsymbol{z}^{\text{f}}$ and the text embedding $\boldsymbol{t}_i$.
The final visual-text bidirectional contrastive learning loss $\mathcal{L}_{\text{vtc}}$ is the average of the sum of the visual-to-text loss $\mathcal{L}_{\text{v2t}}$ and the text-to-visual loss $\mathcal{L}_{\text{t2v}}$, which can be expressed as:
\begin{equation}
\begin{gathered}
    \mathcal{L}_{\mathrm{v2t}}=-\sum_{i=1}^K\log\frac{\exp(\cos(\boldsymbol{z^\text{f}}_{ij},\boldsymbol{t}_i)/\tau)}{\sum_{j=1}^B\exp(\cos(\boldsymbol{z^\text{f}}_{ij},\boldsymbol{t}_j)/\tau)}, 
    \\
    \mathcal{L}_{\mathrm{t2v}}=-\sum_{i=1}^{B}\log\frac{\exp(\cos(\boldsymbol{z^\text{f}}_{ji},\boldsymbol{t}_{i})/\tau)}{\sum_{j=1}^{K}\exp(\cos(\boldsymbol{z^\text{f}}_{ji},\boldsymbol{t}_i)/\tau)},
    \end{gathered}
\end{equation}
where $\cos(\cdot,\cdot)$ is the cosine similarity and $\tau$ is the temperature parameter.
\begin{figure}
    \centering
    \includegraphics[width=1\linewidth]{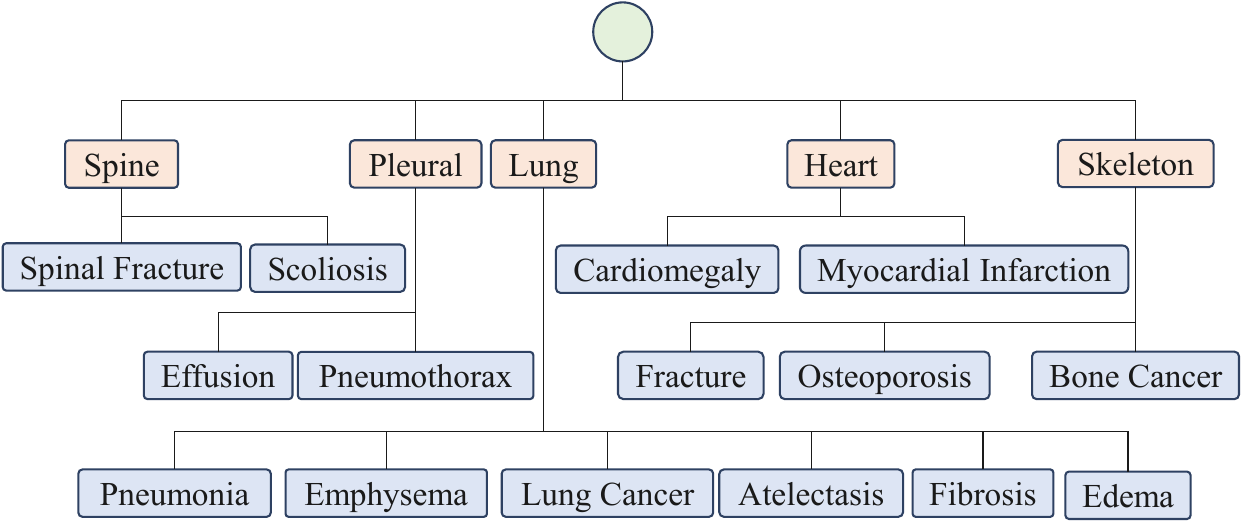}
    \caption{Preconstruct a knowledge graph. In the graph, the green circle represent a global node, orange boxes represent organ-level entities, and blue boxes represent key findings.}
    \label{fig:graph}
\end{figure}

\subsection{Knowledge graph embedding module}
To enhance the reasoning ability in Med-VQA, we propose a knowledge graph embedding (KGE) module to effectively integrate structured medical knowledge into the model. As shown in Fig.~\ref{fig:framework_new}(b), the KGE module utilizes a GAT, which leverages a pre-constructed knowledge graph to assign varying weights to the neighbors of each node through the attention mechanism. GAT facilitates the weighted aggregation of features from important nodes, embedding them into the model's feature space. For each knowledge graph node, we compute its attention weight, and then the structural and semantic information from the knowledge graph is embedded into a low-dimensional vector space.

While FCFA aligns explicit visual-textual features, it is often limited by surface-level co-occurrences in the training data. KGE complements this by injecting knowledge-driven constraints through a GAT-encoded medical topology. By capturing implicit structural dependencies (e.g., organ-lesion relations) absent from image-text pairs, KGE enables out-of-distribution reasoning. Consequently, this integration transforms simple associative matching into logical clinical reasoning, ensuring that the model adheres to established anatomical rules.

Specifically, the knowledge graph $k$ proposed in SLAKE~\cite{liu2021slake} is used to highlight disease keywords and strengthen the relationships between them. An adjacency matrix $\boldsymbol{M}_{\text{adj}}=\{e_{ij}\} \in\mathbb{R}^{N_v \times N_v}$, where $N_v$ denotes the total number of nodes in the pre-constructed knowledge graph, is constructed to represent the edges $e$ based on the relationships between diseases and organs. Each node represents a disease keyword, and we set $e_{ij}$ to 1 when the source node \(n_i\) is connected to the target node $n_j$. Nodes associated with the same organ or tissue are connected to each other and to the root node, as shown in Fig.~\ref{fig:graph}. 

After obtaining the adjacency matrix $\boldsymbol{M}_{\text{adj}}$, binary values (0 and 1) are used to represent the relationships between organs and diseases. The organ-disease features $\boldsymbol{f}_{\text{od}}$ are tokenized and embedded via RoBERTa.
These features are then passed into a single-layer GAT module to extract valuable information about organ-disease relationships, denoted as $ \boldsymbol{f}_{\text{g}}\in\mathbb{R}^{N_v \times d} $, which can be summarized as follows:
\begin{equation}
   \boldsymbol{f}_{\text{g}} = \text{GAT}(\boldsymbol{f}_{\text{od}}, \boldsymbol{M}_{\text{adj}}),
\end{equation}
where $\boldsymbol{f}_{\text{od}}\in\mathbb{R}^{N_v \times d} $ represents the input organ-disease node features, and 
$d$ denotes the embedding dimension.

We employ $ \boldsymbol{f}_{\text{g}} $ as the Query and $\boldsymbol{t}$ as Key/Value. By doing so, we constrain the output $\boldsymbol{k}'$ to the semantic space of the knowledge graph, but with specific nodes highlighted based on their relevance to the question. By fusing the knowledge graph information $ \boldsymbol{f}_{\text{g}} $ with images and texts (questions) via CrossAttention, we obtain the final representation $\boldsymbol{k}'\in\mathbb{R}^{N_v \times d} $:
\begin{equation}
    \begin{aligned}
    \boldsymbol{k}' &= \text{CrossAttention}( \boldsymbol{f}_{\text{g}}, \boldsymbol{t}) \\
   & = \text{Softmax}\left( \frac{(\boldsymbol{f}_{\text{g}} W''_Q)(\boldsymbol{t} W''_K)^\intercal}{\sqrt{d}} \right) (\boldsymbol{t} W''_V),
    \end{aligned}
\end{equation}
where $W''_Q, W''_K, W''_V \in \mathbb{R}^{d \times d}$ are learnable parameters for the cross-attention in Fig.~\ref{fig:framework_new}(b).

To this end, the KGE module effectively integrates structured medical knowledge into the model, offering rich prior knowledge support for subsequent cross-modal interactions, thereby enhancing the model's ability to reason about medical images and questions.

\begin{figure}
    \centering
    \includegraphics[width=0.8\linewidth]{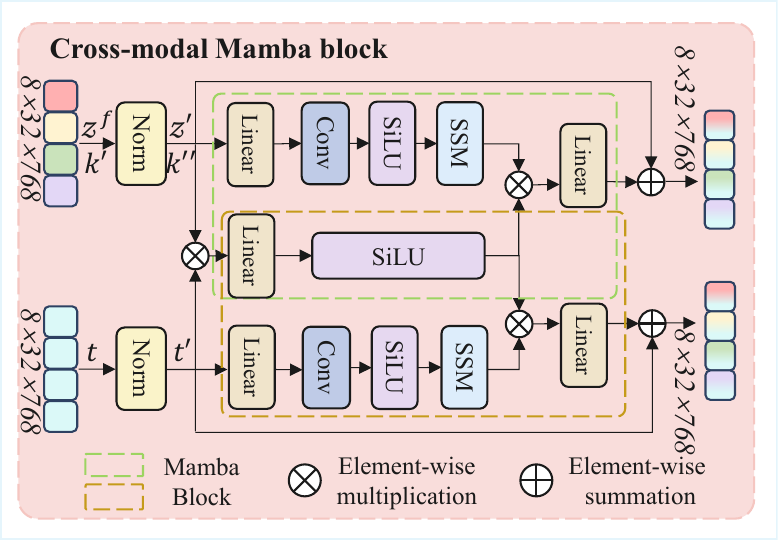}
    \caption{The detailed structure of the cross-modal Mamba (CMM) block.}
    \label{fig:module 2}
\end{figure}
\subsection{Cross-modal interaction representation module}
While Mamba-based methods offer significant computational advantages over Transformer-based multimodal fusion approaches, the sequential scanning mechanism of Mamba limits its ability to effectively learn cross-modal correspondences. To address this, we propose cross-modal Mamba (CMM) blocks (Fig.~\ref{fig:module 2}) within the CMIR module (Fig.~\ref{fig:framework_new}(c)). The CMM blocks maintain the sequential nature of Mamba and its linear computational complexity, while enabling effective cross-modal interactions among learned visual representations, text embeddings, and prior knowledge graph features.

Unlike standard cross-attention which suffers from quadratic complexity $O(L^2)$ due to the global $QK^\top$ computation, CMM leverages the selective scan mechanism of Mamba to achieve linear complexity $O(L)$. This is particularly advantageous in medical tripartite interaction, where the combined sequence of visual tokens, textual embeddings, and knowledge graph features can be long. The CMM selectively propagates relevant information through the hidden state, effectively filtering out modality-specific noise while capturing deep cross-modal correlations.

Given the normalized learnable queries $\boldsymbol{z}^{'}$, the normalized text embeddings $\boldsymbol{t}^{'}$, and the knowledge graph information $\boldsymbol{k}^{''}$, we construct two pairs of multimodal feature sequences, ($\boldsymbol{z}^{\text{t2z}}$, $\boldsymbol{t}^{\text{z2t}}$) and ($\boldsymbol{k}^{\text{t2k}}$, $\boldsymbol{t}^{\text{k2t}}$), by interleaving features from different modalities:
\begin{equation}
\begin{gathered}
\boldsymbol{z}^{\text{t2z}}=\text{Fus}(\text{Mamba}(\boldsymbol{z}^{'}, \boldsymbol{z}^{'} \otimes \boldsymbol{t}^{'})) \oplus \boldsymbol{z}^{'},\\
\boldsymbol{t}^{\text{z2t}}=\text{Fus}(\text{Mamba}(\boldsymbol{t}^{'}, \boldsymbol{t}^{'} \otimes \boldsymbol{z}^{'})) \oplus \boldsymbol{t}^{'},
\end{gathered}
\end{equation}

\begin{equation}
\begin{gathered}
\boldsymbol{k}^{\text{t2k}}=\text{Fus}(\text{Mamba}(\boldsymbol{k}^{''}, \boldsymbol{k}^{''} \otimes \boldsymbol{t}^{'})) \oplus \boldsymbol{k}^{''},\\
\boldsymbol{t}^{\text{k2t}}=\text{Fus}(\text{Mamba}(\boldsymbol{t}^{'}, \boldsymbol{t}^{'} \otimes \boldsymbol{k}^{''})) \oplus \boldsymbol{t}^{'},
\end{gathered}
\end{equation}
where $\text{Mamba}(\cdot,\cdot)$ denotes the original Mamba function, $\text{Fus}(\cdot,\cdot)$ represents the fusion function, $\otimes$ and $\oplus$ represent element-wise multiplication and element-wise summation, respectively. It is noted that the $\text{Fus}(\cdot,\cdot)$ function is implemented as a linear transformation in our work.

The cross-Mamba interaction representations for downstream tasks are obtained after two sets of CMM blocks ($\boldsymbol{Num}$ = 2). This number of blocks is empirically determined to provide sufficient cross-modal reasoning capacity while preventing over-fitting on the relatively small-scale medical datasets. Detailed experimental analysis regarding the choice of $\boldsymbol{Num}$ is provided in Section~\ref{sec:cmm_block_num}. We treat $\boldsymbol{t}^{\text{z2t}}$ (text fused with visual features) as the primary representation. Hyperparameters $\eta$ and $\theta$ regulate the contribution of the supplementary knowledge graph features relative to this primary baseline. Finally, we concatenate these features after two sets of CMM blocks to obtain high-representative fused features, denoted as $\boldsymbol{x}^{\text{f}}$:
\begin{equation}
\begin{gathered}
\boldsymbol{x}^{\text{text}}=\boldsymbol{t}^{\text{z2t}} + \eta\boldsymbol{t}^{\text{k2t}} + \theta\boldsymbol{k}^{\text{t2k}},\\
\boldsymbol{x}^{\text{f}}=\text{Concat}\left(\boldsymbol{z}^{\text{t2z}} , \boldsymbol{x}^{\text{text}} \right),
\end{gathered}
\end{equation}
where $\eta$ and $\theta$ are the weight factors, $\text{Concat}$ denotes the concatenation operation, $\boldsymbol{z}^{\text{t2z}}$ is the visual feature sequence enriched with textual information, and $\boldsymbol{x}^{\text{text}}$ is the aggregated textual feature enriched with both visual and knowledge graph information. The CMM enables a selective scanning mechanism in Mamba to capture both intra- and inter-modal dependencies.

\begin{figure}
    \centering
    \includegraphics[width=0.6\linewidth]{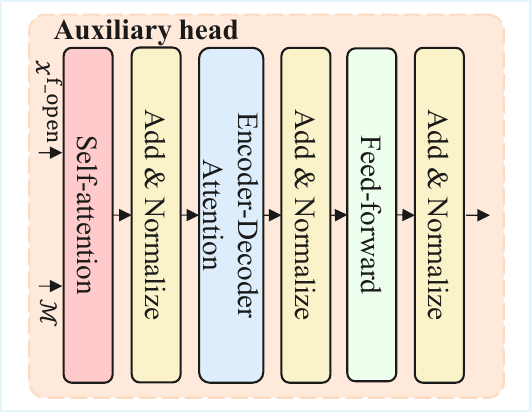}
    \caption{The detailed structure of the Auxiliary head (AHead).}
    \label{fig:module 3}
\end{figure}
\subsection{Free-form answer enhanced multi-task learning module }
Traditional Med-VQA methods primarily focus on closed-ended questions, where the model predicts answers from a predefined set of options. However, in clinical practice, open-ended questions are common, requiring the model to generate free-form answers based on the visual and textual information provided. Thus, we hypothesize that generating answers for open-ended questions is significantly more demanding than simply predicting a ``yes'' or ``no'' response. Strict penalization to encourage the generation of more precise free-form answers is expected to improve the model's overall performance.

Based on the hypothesis, as shown in Fig.~\ref{fig:framework_new}(d), we propose a free-form answer enhanced multi-task learning (FAMT) module, which consists of a classification head and an auxiliary head. During training, we utilize the question type annotations provided by the datasets. The auxiliary loss is computed only for indices $i$ where the question type is open-ended, effectively masking closed-ended questions from this specific module.

\textbf{Classification head.} The classification head is designed to handle traditional classification tasks, including both closed-ended and open-ended question classification. It consists of a set of linear layers, a normalization layer, and a GELU layer. The fused feature vector, $\boldsymbol{x}^{\text{f}}$, is classified using a binary cross-entropy loss function, denoted as $\mathcal{L}_{\text{cls}}$.

\textbf{Auxiliary head.} As shown in Fig.~\ref{fig:module 3}, the auxiliary head (AHead) focuses on generating free-form answers and adopts a pre-trained T5 decoder~\cite{raffel2020exploring} as its basic architecture. To ensure that the model focuses on information relevant to the question, a learnable attention mask $\mathcal{M}$ is introduced, which can automatically select important tokens from multimodal features. Specifically, we first filter out all the close-ended embeddings in the batch, obtaining the pure open-ended question embeddings $\{\boldsymbol{x}_{i}^{\text{f\_open}}\}_{i=1}^{O}$ with corresponding answers $\{a_i\}_{i=1}^{O}$, where $O$ is the number of open-ended samples in the batch ($O \leq B$). The T5 decoder generates corresponding answer prediction $\hat{a}_i$ for $\boldsymbol{x}_{i}^{\text{f\_open}}$. 
Next, we initialize the mask $\mathcal{M}$, with all elements initially set to 1. The learnable mask guides the model to focus on relevant tokens within the multi-modal features of open-ended questions.
Finally, we use mask-guided cross-entropy loss $\mathcal{L}_{\text{ce}}$ as our objective function:
\begin{equation}    
\mathcal{L}_{\text{aux}} = \mathcal{L}_{\text{ce}}(\hat{a}_i, \mathcal{M}, a_i),
\end{equation}

To this end, the overall loss function $\mathcal{L}$ for multi-task learning is composed of the classification loss $\mathcal{L}_\text{cls}$, the cross-modal contrastive loss $\mathcal{L}_\text{vtc}$, and the auxiliary task loss $\mathcal{L}_\text{aux}$, and is expressed as:
\begin{equation}
    \mathcal{L} = \mathcal{L}_{\text{cls}} + \alpha \mathcal{L}_{\text{vtc}} + \beta \mathcal{L}_{\text{aux}}.
\end{equation}
where $\alpha$ and $\beta$ are the weight factors. By jointly optimizing these loss functions, the model can simultaneously enhance classification accuracy and the quality of free-form answer generation. The auxiliary free-form answer can then benefit the classification performance, thereby enabling better handling of various types of medical questions. Note that the generative Auxiliary Head is employed solely during the training phase to enhance representation learning. During inference, the final answer is predicted via the Classification Head.

\section{Experiments}
\subsection{Datasets and setting}
\textbf{Dataset and metrics.} The \textbf{SLAKE}~\cite{liu2021slake} dataset supports both Chinese and English languages, covering a wide variety of organ types. In this study, we focus on the English subset. SLAKE is partitioned at the image level, with 70\% (450 images) used for training, 15\% (96 images) for validation, and 15\% (96 images) for testing, following the approach in~\cite{chen2024mapping}. The \textbf{VQA-RAD}~\cite{lau2018dataset} dataset comprises radiological images annotated by volunteers with professional medical expertise. It contains 315 radiological images from three organ types and 3,515 clinical questions. To ensure fair comparisons with previous studies, we follow the established partitioning~\cite{li2023masked}, allocating 3,064 questions for training and 451 for testing. The \textbf{OVQA}~\cite{huang2022ovqa} dataset specializes in orthopedics, consisting of 19,020 medical visual question-answer pairs derived from 2,001 medical images. 
Following the original paper's protocol~\cite{van2023open}, we split OVQA into training (80\%, 15,216 questions), validation (10\%, 1,902 questions), and testing (10\%, 1,902 questions). The performance is evaluated using the accuracy (ACC) metric.

{\textbf{Implementation details.}} Our method is implemented in PyTorch and executed on a single NVIDIA GeForce RTX 4090 GPU. The model is optimized using the AdamW optimizer with an initial learning rate of $5 \times 10^{-6}$. Training is conducted for 50 epochs with a batch size of 8. The hyperparameters $\alpha$, $\beta$, $\eta$, and $\theta$ are empirically set to 0.2, 0.3, 0.1, and 0.1, respectively. Medical images are resized to 384 $\times$ 384. The dimension of the learnable query vector is set to 32. To ensure feature consistency with the backbones, the hidden embedding dimension, denoted as $d$, is set to 768 for both the cross-modal interaction modules and the Mamba blocks.

\begin{table*}
\centering
\caption{Performance comparison (ACC, \%) with state-of-the-art methods on the SLAKE, VQA-RAD, and OVQA datasets. Results for methods without an asterisk are reported as Mean $\pm$ Standard Deviation. The best values are in \textbf{bold}. Improvements over the \underline{second-best} results are highlighted in {\color{red}red}.}\label{tab:comparison}

\renewcommand\tabcolsep{4pt} 
\renewcommand{\arraystretch}{1.3}

\resizebox{\textwidth}{!}{
    \begin{tabular}{c|ccc|ccc|ccc}
    \hline
        \multirow{2}{*}{Method}&   \multicolumn{3}{c|}{SLAKE}&\multicolumn{3}{c|}{VQA-RAD}&\multicolumn{3}{c}{OVQA}\\ \cline{2-10}
        &  Open & Closed & Overall & Open & Closed & Overall & Open & Closed & Overall \\
    \hline
    M2I2~\cite{li2023self} & \ms{73.21}{0.25} & \ms{89.50}{0.12} & \ms{77.60}{0.18} & \ms{48.78}{0.30} & \ms{82.71}{0.15} & \ms{69.16}{0.21} & \ms{36.02}{0.45} & \ms{72.38}{0.32} & \ms{57.89}{0.28} \\
    MUMC~\cite{li2023masked} & \ms{75.04}{0.22} & \ms{93.27}{0.10} & \ms{82.19}{0.15} & \ms{62.57}{0.28} & \ms{84.19}{0.14} & \ms{75.61}{0.19} & \ms{39.05}{0.42} & \ms{75.87}{0.30} & \ms{61.20}{0.25} \\
    M3AE~\cite{chen2024mapping} & \ms{80.78}{0.18} & \ms{87.65}{0.14} & \ms{83.47}{0.12} & \ms{63.50}{0.25} & \ms{85.42}{0.11} & \ms{76.72}{0.16} & \underline{\ms{61.65}{0.38}} & \underline{\ms{84.37}{0.15}} & \underline{\ms{75.32}{0.22}} \\
    
    UnICLAM$^{*}$~\cite{UnICLAM} & 81.10 & 85.70 & 83.10 & 59.80 & 82.60 & 73.20 & - & - & - \\
    MEVF$^{*}$~\cite{nguyen2019overcoming} & - & - & - & 37.41 & 75.60 & 60.42 & 34.69 & 74.21 & 58.46 \\
    MPR~\cite{ossowski2023retrieving} & \ms{75.50}{0.28} & \ms{81.50}{0.22} & \ms{77.90}{0.24} & \ms{59.30}{0.32} & \ms{79.10}{0.20} & \ms{73.30}{0.25} & \ms{31.70}{0.48} & \ms{72.60}{0.35} & \ms{56.97}{0.30} \\
    
    LaPA~\cite{gu2024lapa} & \ms{80.00}{0.15} & \ms{87.25}{0.18} & \ms{82.84}{0.14} & \ms{67.87}{0.22} & \ms{84.92}{0.12} & \underline{\ms{78.15}{0.15}} & \ms{56.46}{0.35} & \ms{83.12}{0.18} & \ms{72.51}{0.20} \\
    
    MMQ$^{*}$~\cite{do2021multiple} & - & - & - & 52.00 & 76.71 & 66.92 & 46.04 & 76.13 & 64.14 \\
    MQAT~\cite{liu2022transformer} & \ms{77.20}{0.24} & \ms{87.97}{0.15} & \ms{81.50}{0.18} & \ms{48.60}{0.35} & \ms{77.11}{0.18} & \ms{65.50}{0.22} & \ms{38.21}{0.40} & \ms{79.92}{0.28} & \ms{67.00}{0.32} \\
    VQAMix$^{*}$~\cite{gong2022vqamix} & - & - & - & 56.90 & 79.50 & 70.50 & 44.98 & 76.83 & 64.14 \\
    
    VG-CALF~\cite{VG-CALF} & \underline{\ms{81.64}{0.12}} & \ms{84.23}{0.15} & \ms{83.43}{0.10} & \ms{67.47}{0.20} & \ms{85.28}{0.14} & \ms{76.16}{0.18} & \ms{60.74}{0.32} & \ms{83.96}{0.12} & \ms{74.21}{0.24} \\
    CCIS-MVQA$^{*}$~\cite{CCIS-MVQA} & 80.12 & 86.72 & \underline{84.08} & \textbf{68.78} & 79.24 & 75.06 & - & - & - \\
    \hline
    KG-CMI (Ours) & \textbf{\ms{82.78}{0.08}}$_{\color{red}{\uparrow 1.14}}$ & \ms{87.47}{0.05} & \textbf{\ms{84.26}{0.06}}$_{\color{red}{\uparrow 0.18}}$ & \underline{\ms{68.18}{0.10}}$_{\downarrow 0.6}$  & \ms{84.11}{0.08} & \textbf{\ms{78.21}{0.09}}$_{\color{red}{\uparrow 0.06}}$ & \textbf{\ms{71.77}{0.12}}$_{\color{red}{\uparrow 10.12}}$ & \textbf{\ms{84.47}{0.10}}$_{\color{red}{\uparrow 0.10}}$ & \textbf{\ms{79.58}{0.11}}$_{\color{red}{\uparrow 4.26}}$ \\
    \hline
    \end{tabular}
}
\end{table*}

\begin{table*}[h!]
\centering
\caption{Performances of ablation study on multiple datasets. QQ-F denotes the QQ-Former module.}
\label{tab:ablation}
\renewcommand\arraystretch{1.3} 
\begin{tabular}{c|ccccc|ccc|ccc|ccc}
\hline
\multirow{2}{*}{Type} &\multirow{2}{*}{QQ-F} & \multirow{2}{*}{CMCL} & \multirow{2}{*}{KGE} & \multirow{2}{*}{CMM} & \multirow{2}{*}{AHead} & \multicolumn{3}{c|}{SLAKE} & \multicolumn{3}{c|}{VQA-RAD} & \multicolumn{3}{c}{OVQA} \\ \cline{7-15}
 & & & & & & Open & Closed & Overall & Open & Closed & Overall & Open & Closed & Overall \\
\hline
 Baseline & &  &  &  & & 80.31 & 85.25 & 82.25 & 60.52 & 78.31 & 71.25 & 61.65 & 84.37 & 75.32 \\
 
 Addition & $\checkmark$ &  &  &  &  & 80.45 & 86.20 & 82.58 & 65.34 & 79.78 & 74.11 & 68.95 & 83.56 & 77.74 \\

 & & $\checkmark$ &  &  & & 80.72 & 85.96 & 82.77 & 64.77 & 80.51 & 74.33 & 66.09 & 84.42 & 77.07 \\
 
  &  & & $\checkmark$ &  & & 81.78 & 85.47 & 83.12 & 68.11 & 83.25 & 77.11 & 69.11 & 84.24 & 78.48 \\
  
 & &  &  & $\checkmark$ & & 81.49 & 85.96 & 83.24 & 65.34 & 79.41 & 73.88 & 68.07 & 83.74 & 77.48 \\
 
 & &  &  &  & $\checkmark$ & 81.34 & 84.02 & 82.39 & 66.47 & 77.94 & 73.43 & 68.73 & 82.86 & 77.22 \\

Ablation & $\checkmark$ & $\checkmark$ &  & $\checkmark$ & $\checkmark$ & 81.95 & 86.84 & 83.72 &	67.05 & 82.15 & 77.42 & 70.25 & 83.35 & 78.85 \\

& $\checkmark$ & $\checkmark$ & $\checkmark$ & $\checkmark$ &  & 81.12 & 87.25 & 83.51 & 65.45 & 83.95 & 77.05 & 68.85 & 84.25 & 78.32 \\

Full & $\checkmark$ & $\checkmark$ & $\checkmark$ & $\checkmark$ & $\checkmark$ & 82.78 & 87.47 & 84.26 & 68.18 & 84.11 & 78.21 & 71.77 & 84.47 & 79.58 \\
\hline
\end{tabular}
\end{table*}

\subsection{Experimental results}
\subsubsection{Comparison with the state-of-the-art methods}
To evaluate the effectiveness of KG-CMI, we compare it against several state-of-the-art methods. These include: (1) Pre-training and downstream task-based methods, such as M2I2~\cite{li2023self}, MUMC~\cite{li2023masked}, M3AE~\cite{chen2024mapping}, and UniCLAM~\cite{UnICLAM}; (2) Knowledge-enhanced methods, including MEVF~\cite{nguyen2019overcoming}, MPR~\cite{ossowski2023retrieving}, and LaPA~\cite{gu2024lapa}; and (3) Architectural-specialized methods such as MMQ~\cite{do2021multiple}, MQAT~\cite{liu2022transformer}, VQAMix~\cite{gong2022vqamix}, VG-CALF~\cite{VG-CALF}, and CCIS-MVQA~\cite{CCIS-MVQA}. The comparative results are summarized in Table~\ref{tab:comparison}, which reports the accuracy for Open-ended questions, Closed-ended questions, and the Overall performance across the SLAKE, VQA-RAD, and OVQA datasets. Note that the results marked with an asterisk (*) in Table \ref{tab:comparison} are directly cited from the original papers since the corresponding official codes are unavailable, while all other results are obtained by running the official codes under the same experimental settings.

To ensure the reliability of the results, all experiments for KG-CMI are conducted over five independent runs with different random seeds. As shown in Table~\ref{tab:comparison}, we report the average accuracy along with the standard deviation. To verify that the improvements are not due to random fluctuations, we performed a two-tailed t-test comparing KG-CMI with the second-best baseline on each dataset. The resulting p-values (all $<$ 0.05) indicate that our framework’s gains are statistically significant, providing a robust improvement over existing state-of-the-art methods.

Specifically, on the SLAKE dataset, KG-CMI achieves an overall accuracy of 84.26\%, surpassing the second-best method CCIS-MVQA (84.08\%) with a 0.18\% improvement. On the VQA-RAD dataset, KG-CMI attains an overall accuracy of 78.21\%, ahead of LaPA (78.15\%) with an improvement of 0.06\%. On the OVQA dataset, KG-CMI achieves 79.58\%, substantially outperforming M3AE (75.32\%, with an improvement of 4.26\%) and LaPA (72.51\%).

Notably, KG-CMI achieves the best performance in open-ended question answering tasks across all datasets. On SLAKE, it scores 82.78\%, a 1.14\% improvement over VG-CALF (81.64\%, which ranks second). On VQA-RAD, it scores 68.18\%, although this is a slight decrease of 0.6\% compared to CCIS-MVQA (68.78\%), but it still leads overall. On OVQA, it scores 71.77\%, representing a 10.12\% improvement over M3AE (61.65\%).
This performance is largely attributed to the auxiliary FAMT module, which enhances the model's ability to handle open-ended Med-VQA tasks and boosts robustness in addressing various medical questions. Additionally, the integration of the knowledge graph has improved the model's performance in closed-ended tasks. The coordinated interaction between these components, including FCFA, KGE, CMIR, and FAMT, enables KG-CMI to outperform other methods in both closed-ended and open-ended scenarios, underscoring the value of this multi-module approach.

\begin{figure*}[h!]
    \centering
    \includegraphics[width=1\linewidth]{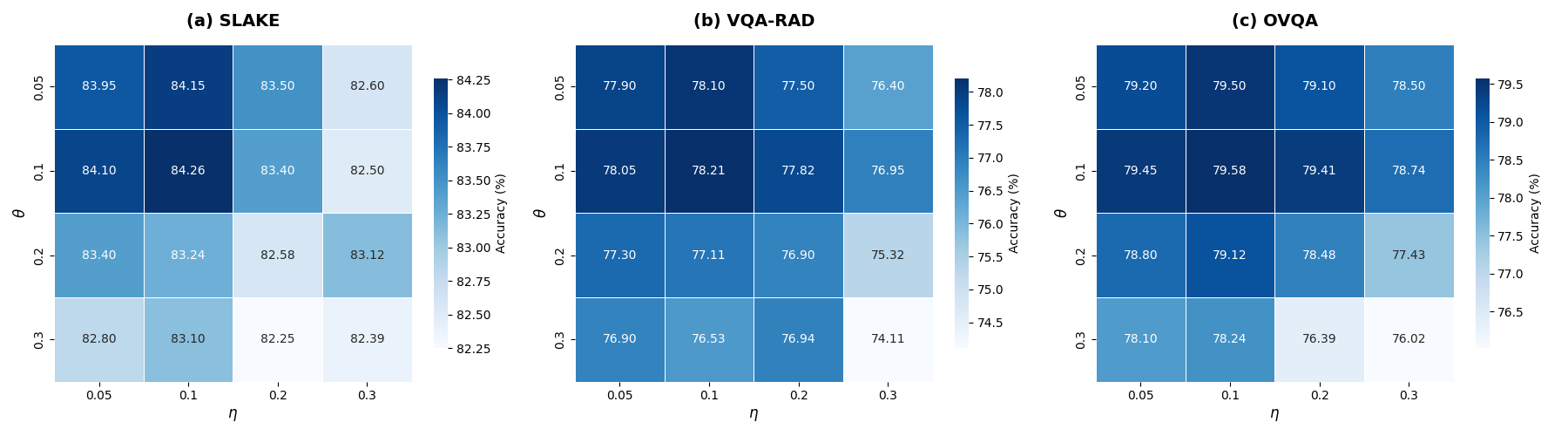}
    \caption{Performance comparisons of different $\eta$ and $\theta$ values on three datasets.}
    \label{fig:para_ablation}
\end{figure*}

\subsubsection{Ablation study} To assess the impact of different components in the KG-CMI framework, we conduct an ablation study on three benchmark datasets. The study systematically evaluates the model's performance by individually adding key modules: QQ-Former, CMCL, KGE, CMM, and AHead.
The results, shown in Table~\ref{tab:ablation}, demonstrate that the inclusion of each module improves the overall performance across all datasets. Specifically, when QQ-Former, CMCL, KGE, and CMM are incorporated, the model shows a significant improvement in both open-ended and closed-ended tasks, as well as overall performance.

Specifically, QQ-Former significantly improves the performance of open-ended tasks, with SLAKE increasing to 80.45\%, VQA-RAD reaching 65.34\%, and OVQA improving to 68.95\%. CMCL enhances closed-ended task accuracy, with SLAKE rising to 85.96\% and VQA-RAD reaching 80.51\%. KGE greatly benefits medical knowledge-intensive datasets such as SLAKE and VQA-RAD. CMM enhances the model's adaptability to complex scenarios, with the overall performance on SLAKE reaching 83.24\%. AHead provides noticeable gains in open-ended scenarios, increasing the scores of SLAKE, VQA-RAD, and OVQA by 1.03\%, 5.95\%, and 7.08\%, respectively. When all modules are combined, the performances of SLAKE, VQA-RAD, and OVQA reach their highest values. The results fully demonstrate that, under the synergistic effect of each module, the model achieves optimal performance across all evaluated tasks, further verifying the performance gains attributable to each individual module.

To further validate the necessity of each component, we perform a series of ablation experiments by excluding individual modules. As shown in the last few rows of Table~\ref{tab:ablation}, removing the AHead module results in a notable decrease in accuracy for open-ended questions. This is because AHead, acting as a generative auxiliary task during training, forces the model to learn more robust and fine-grained multimodal representations. Similarly, removing the KGE module leads to a drop in overall performance (e.g., -0.54\% on SLAKE), highlighting the critical role of structured medical knowledge in bridging the gap between visual symptoms and clinical concepts. Overall, the ablation study confirms that the modules play complementary roles, and their full integration is necessary for KG-CMI to achieve its peak predictive accuracy.

\begin{table}[h]
\centering
\caption{Performance comparison of open-ended answers using NLG metrics.}\label{tab:nlg_comparison}
\setlength{\tabcolsep}{5pt}
\renewcommand{\arraystretch}{1.3}
\begin{tabular}{c|c|cccc}
\hline
Dataset & Method & BLEU-1 & BLEU-4 & METEOR & CIDEr \\ \hline
\multirow{3}{*}{SLAKE} & MUMC~\cite{li2023masked} & 0.421 & 0.285 & 0.242 & 1.125 \\
 & M3AE~\cite{chen2024mapping} & 0.452 & 0.312 & 0.265 & 1.218 \\
 & \textbf{KG-CMI} & \textbf{0.485} & \textbf{0.342} & \textbf{0.291} & \textbf{1.424} \\ \hline
\multirow{3}{*}{VQA-RAD} & MUMC~\cite{li2023masked} & 0.385 & 0.212 & 0.195 & 0.852 \\
 & M3AE~\cite{chen2024mapping} & 0.402 & 0.235 & 0.210 & 0.925 \\
 & \textbf{KG-CMI} & \textbf{0.438} & \textbf{0.268} & \textbf{0.235} & \textbf{1.085} \\ \hline
\end{tabular}
\end{table}

\begin{table}[ht]
    \renewcommand\arraystretch{1.3}
    \setlength{\tabcolsep}{2.7mm}
    \centering
    \caption{Comparisons of different $\alpha$ and $\beta$ values on the VQA-RAD dataset.}\label{tab:ablation_both}
    \setlength{\tabcolsep}{2.0mm}
        \begin{tabular}{c|ccc||c|ccc}
            \hline
            $\alpha$ & Open & Closed & Overall & $\beta$ & Open & Closed & Overall \\
            \hline
            0.1 & 67.87 & 82.10 & 77.62 & 0.1 & 67.05 & 80.32 & 75.75 \\
            \textbf{0.2} & \textbf{68.18} & \textbf{84.11} & \textbf{78.21} & 0.2 & 67.87 & 82.10 & 77.62 \\
            0.3 & 66.50 & 83.77 & 76.24 & \textbf{0.3} & \textbf{68.18} & \textbf{84.11} & \textbf{78.21} \\
            0.4 & 68.02 & 79.84 & 75.51 & 0.4 & 66.50 & 83.77 & 76.24 \\
            0.5 & 67.34 & 78.64 & 73.89 & 0.5 & 68.02 & 80.05 & 75.92 \\
            \hline
        \end{tabular}
\end{table}

\subsubsection{Evaluation of open-ended answer generation}
To further evaluate the quality of the open-ended answers generated by KG-CMI, we employ standard natural language generation evaluation metrics, including BLEU~\cite{papineni2002bleu}, METEOR~\cite{banerjee2005meteor}, and CIDEr~\cite{vedantam2015cider}. Unlike accuracy, which requires an exact match, these metrics evaluate the fluency and semantic similarity of the output. As presented in Table~\ref{tab:nlg_comparison}, KG-CMI consistently outperforms generative baselines across all metrics. For instance, on the SLAKE dataset, KG-CMI achieves a CIDEr score of 1.42, a significant improvement over MUMC. This performance gain underscores AHead’s capacity to capture complex clinical semantics and translate them into more natural and accurate linguistic responses.

\subsection{Parametric sensitivity and qualitative evaluation}

\subsubsection{Effectiveness of weight factors}
We explore the effectiveness of weight factors $\alpha$ and $\beta$ in multi-task learning. We fix the coefficient of the primary classification loss  $\mathcal{L}_\text{cls}$ to 1.0 to serve as an anchor, optimizing the relative contributions of the auxiliary losses $\alpha$ and $\beta$. To identify the optimal settings, we conduct a coordinate-wise sensitivity analysis for $\alpha$ and $\beta$. Table~\ref{tab:ablation_both} summarizes the model’s performance by individually varying one weight factor while fixing the other at its identified empirical optimum (e.g., $\beta=0.3$ when evaluating $\alpha$, and vice versa). As shown in Table~\ref{tab:ablation_both}, performance declines significantly when the weight factors $\alpha$ and $\beta$ exceed 0.2 and 0.3, respectively. This decline is likely due to the increased influence of the auxiliary head and contrastive loss affecting the main branch. When $\alpha$ and $\beta$ exceed 0.2 and 0.3 respectively, the positive impact of these auxiliary tasks becomes negative, disrupting the stable feature representation learned by the main branch. Based on these findings, we set the weight factors $\alpha$ and $\beta$ to 0.2 and 0.3 respectively in the experiments.

Additionally, we investigate the role of hyperparameters $\eta$ and $\theta$ in regulating the interaction of cross-modal information, which impacts the accuracy of the final prediction results. Fig.~\ref{fig:para_ablation} uses a set of heatmaps to illustrate the effect of various $\eta$ and $\theta$ values on the fusion of cross-modal features across three benchmark datasets. We systematically vary the hyperparameters $\eta$ and $\theta$ from 0.1 to 0.3, as visualized in the heatmaps in Fig.~\ref{fig:para_ablation}. The results consistently indicate that the combination of $\eta$ = 0.1 and $\theta$ = 0.1 yields the highest accuracy across all three datasets. Additionally, we investigated the performance at a smaller scale (e.g., $\eta$ = 0.05 and $\theta$ = 0.05), observing a slight performance degradation (e.g., -0.21\% on SLAKE). This suggests that when these interaction weights are too low, the model's ability to integrate external medical knowledge and cross-modal features is insufficiently activated, leading to suboptimal reasoning. As $\eta$ and $\theta$ increase beyond 0.1, the performance gradually declines, likely because excessive weights on auxiliary features may introduce noise or overshadow the primary visual-textual representations.

\begin{table}[h]
\renewcommand\arraystretch{1.3}
\setlength{\tabcolsep}{5mm}
\centering
\caption{Performance comparison of different QQ-Former layer depths (results on overall ACC \%).}
\label{tab:qqformer_depth}
\begin{tabular}{l|c|c|c}
\hline
Layers       & SLAKE & VQA-RAD & OVQA  \\
\hline
1 (Default)  & 84.26           & 78.21             & 79.58          \\
2            & 84.38           & 78.05             & 79.62          \\
3            & 84.41           & 77.89             & 79.55          \\
\hline
\end{tabular}
\end{table}

\subsubsection{Sensitivity of QQ-Former layers}
We investigate the impact of the number of stacked layers in QQ-Former on model performance. As shown in Table~\ref{tab:qqformer_depth}, we compare the ACC of KG-CMI with 1, 2, and 3 layers. The results show that increasing the layers to 2 or 3 yields marginal improvements on the SLAKE dataset (e.g., +0.12\% with 2 layers) but leads to a slight decrease on VQA-RAD due to potential over-fitting on smaller datasets. Given the clinical need for efficient inference, we adopt a 1-layer configuration as the default for KG-CMI.

\begin{table}[h]
\renewcommand\arraystretch{1.3}
\setlength{\tabcolsep}{3mm}
\centering
\caption{Impact of CMM Block number ($Num$) variation on performance (Results in Overall ACC \%).}
\label{tab:cmm_blocks}
\begin{tabular}{l|c|c|c}
\hline
 $Num$ & SLAKE & VQA-RAD & OVQA \\
\hline
1                      & 83.52           & 77.45             & 78.10          \\
2 (Default)            & 84.26           & 78.21             & 79.58          \\
3                      & 84.15           & 77.92             & 79.61          \\
4                      & 83.98           & 77.58             & 79.45          \\
\hline
\end{tabular}
\end{table}

\begin{figure*}[h!]
    \centering
    \includegraphics[width=0.75\linewidth]{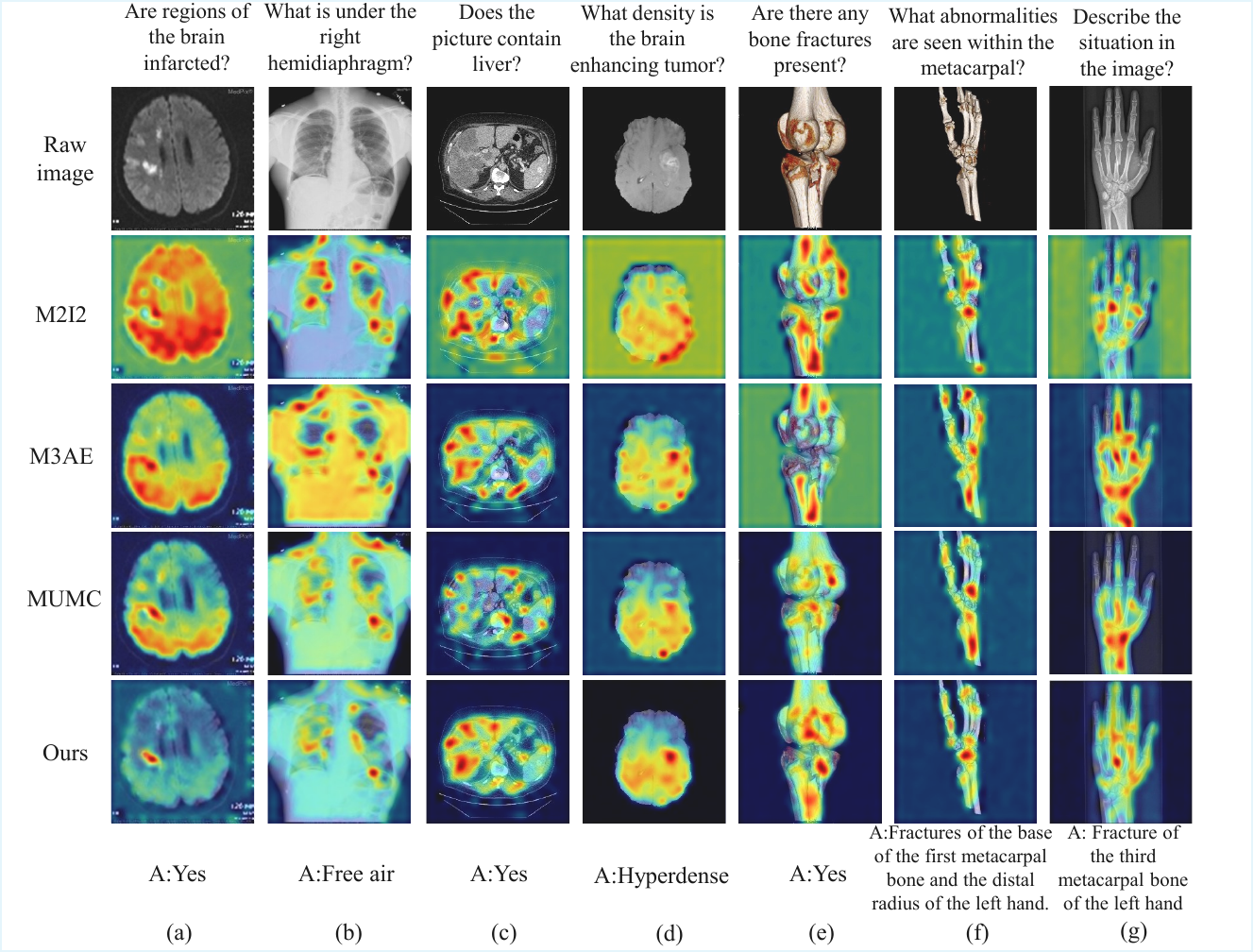}
    \caption{Visual saliency maps for representative methods.}
    \label{fig:visualizations}
\end{figure*}

\subsubsection{Effect of CMM block number variation}
\label{sec:cmm_block_num}
To evaluate the impact of the depth of the CMIR module, we test the KG-CMI framework with different numbers of CMM blocks. As shown in Table~\ref{tab:cmm_blocks}, the performance improves when the number of blocks increases from 1 to 2. Specifically, when the number of blocks is set to 2, the overall accuracy of the model reaches 84.26\%, 78.21\%, and 79.58\% on the SLAKE, VQA-RAD, and OVQA datasets respectively, achieving optimal performance. This indicates that a certain depth is necessary for effective cross-modal interaction. However, further increasing the depth to 3 or 4 results in a performance plateau or a slight decrease in accuracy, particularly on the VQA-RAD dataset. Accordingly, a 2-block configuration strikes the best balance between capturing essential semantic features and maintaining computational efficiency.

\subsubsection{Comparison between CMM and standard cross-attention}
To empirically validate the superiority of CMM, we conduct a comparison with a Transformer-based cross-attention (CA) baseline. As shown in Table~\ref{tab:cmm_ca}, the CMM-based interaction achieves higher ACC with lower FLOPs. Specifically, on the SLAKE dataset, CMM improves the overall ACC by 1.15\% while reducing the FLOPs by approximately 24\%. Such improvements demonstrate that CMM is not only more efficient but also more effective in modeling the complex tripartite interactions in Med-VQA.

\begin{table}[h]
\renewcommand\arraystretch{1.3}
\setlength{\tabcolsep}{2mm}
\centering
\caption{Empirical comparison between CMM and Cross-Attention (CA).}
\label{tab:cmm_ca}
\begin{tabular}{c|c|c|c|c}
\hline
Interaction Module & SLAKE & VQA-RAD & OVQA & FLOPs (G) \\
\hline
Transformer-CA & 83.11 & 77.45 & 75.47 & 18.2 \\
CMM (Ours) & 84.26 & 78.21 & 79.58 & 13.8 \\
\hline
\end{tabular}
\end{table}

\subsubsection{Sensitivity analysis of GAT layers}
Empirical results indicate that a single GAT layer provides the optimal balance for knowledge distillation. As shown in Table~\ref{tab:gatlayer}, since the pre-defined medical KG is restricted to a three-level hierarchy spanning global, organ, and finding nodes, the effective graph diameter is relatively small. Extensive message passing (2-3 layers) tends to dilute the specificity of disease-finding nodes by over-incorporating neighborhood information, essentially creating a `representation bottleneck' where unique clinical features are masked by broad anatomical priors. Thus, a single-hop aggregation is sufficient to capture relevant relational context without compromising node discriminability.

\begin{table}[h]
\renewcommand\arraystretch{1.3}
\setlength{\tabcolsep}{5mm}
\centering
\caption{Sensitivity analysis of GAT layer depth (Results in Overall ACC \%).}
\label{tab:gatlayer}
\begin{tabular}{l|c|c|c}
\hline
GAT Layers       & SLAKE & VQA-RAD & OVQA \\
\hline
1 (Default)  & 84.26 & 78.21 & 79.58 \\
2            & 84.12           & 77.95             & 79.42          \\
3            & 83.89           & 77.58             & 79.21          \\
\hline
\end{tabular}
\end{table}

\subsubsection{Interpretability analysis}
We use Grad-CAM to visualize the attention correlation maps between questions and medical images, and analyze the relevance between the image regions attended by different methods and the questions, as shown in Fig. \ref{fig:visualizations}.

For closed-ended questions, as shown in panels (a), (c), and (e), KG-CMI demonstrates strong localization capabilities by accurately focusing on the target regions directly related to the questions, particularly for clear medical targets such as lesions, organs, and signs. For open-ended questions, as depicted in panels (b) and (d), KG-CMI effectively locates lesion regions in the images through question-based reasoning. In the case of questions with long-sequence answers, such as those in panels (f) and (g), the heatmaps of other methods either fail to identify fine fracture sites or show deviated coverage. In contrast, KG-CMI can precisely localize the positions of fine lesions.

In summary, from the Grad-CAM visualization results (Fig. \ref{fig:visualizations}), it is evident that, compared with M2I2, M3AE, and MUMC, KG-CMI outperforms in focusing on target regions that are directly related to the questions for closed-ended tasks. In open-ended questions, which include both reasoning-type and long-sequence description-type questions, KG-CMI effectively locates lesions, key anatomical regions, and even fine lesions (e.g., metacarpal fractures) by leveraging the combined reasoning of medical knowledge and the question context. These results fully demonstrate that KG-CMI possesses robust reasoning abilities, accurately identifying regions of interest that match the questions and providing essential visual attention support for accurate answers in Med-VQA.

\subsubsection{Hard sample analysis}
\begin{figure*}
    \centering
    \includegraphics[width=0.8\linewidth]{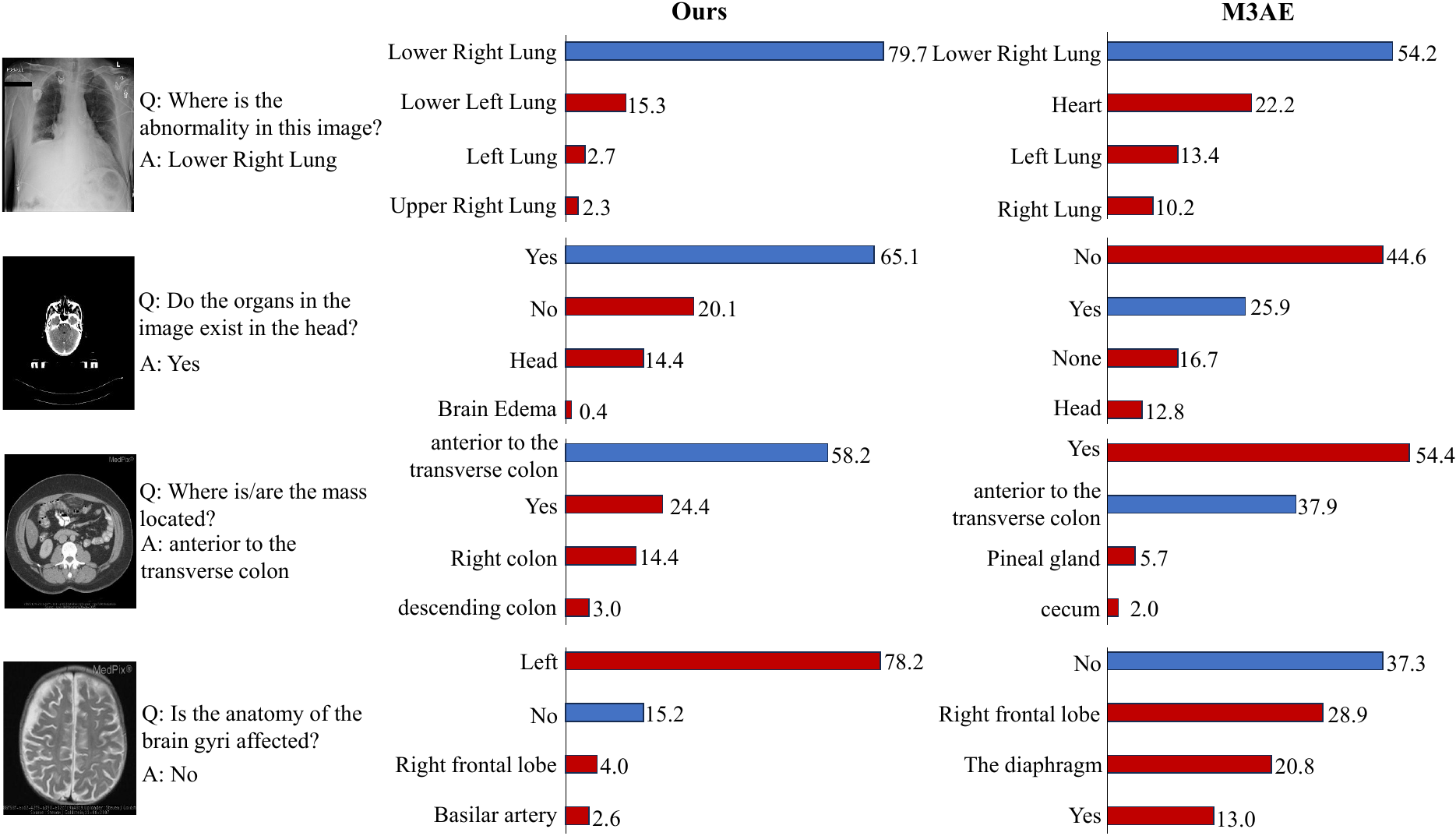}
    \caption{Analysis of failure cases and prediction probability comparison between our approach and M3AE for open-ended and closed-ended questions with answer labels.}
    \label{fig:failure case}
\end{figure*}

To comprehensively assess the strengths and limitations of KG-CMI in Med-VQA, we present four test cases, as illustrated in Fig.~\ref{fig:failure case}. Case 1 involves lung abnormalities, KG-CMI correctly identifies the ``Lower Right Lung'' (79.7\%), whereas M3AE (54.2\%) is confused by neighboring organs like the Heart (22.2\%) and Left Lung (13.4\%). For head organ detection (Case 2), KG-CMI yields 65.1\% for ``Yes,'' while M3AE incorrectly favors ``No'' (44.6\%) and generates irrelevant categories (e.g., ``None''), reflecting judgment reversal. In Case 3 (mass location), KG-CMI achieves 58.2\% accuracy by leveraging external knowledge, while M3AE (37.9\%) incorrectly associates irrelevant regions.

In the erroneous case (Case 4), focusing on disease presence and location, KG-CMI shows a prediction proportion of up to 78.2\% for the incorrect location ``Left'', clearly misjudging the ``affected location''. While KG-CMI can address certain open questions related to medical knowledge, it remains susceptible to issues such as ``location confusion'' and ``existence judgment error'' when localizing lesions in fine anatomical structures and accurately determining disease/organ existence. Although our method fails on this case, it still reveals richer intermediate reasoning than competing models, indicating stronger interpretability and diagnostic insight.

\subsection{Discussion on practical implications}
Beyond numerical accuracy, the practical utility of KG-CMI in clinical settings is evidenced by three key factors. First, the interpretability provided by the attention-aware Grad-CAM (Fig.~\ref{fig:visualizations}) allows clinicians to verify the model's reasoning process, fostering trust in AI-assisted diagnosis. Second, the use of cross-Mamba ensures linear computational complexity, making the model suitable for deployment on standard hospital hardware with limited GPU resources. Third, the framework demonstrates superior robustness and generalization across diverse clinical scenarios. By achieving consistent state-of-the-art performance across three benchmarks (SLAKE, VQA-RAD, and OVQA), KG-CMI proves its efficacy across multiple imaging modalities (e.g., CT, MRI, and X-ray) and varied anatomical regions. Since these datasets originate from different clinical institutions and cover distinct medical specialties such as Radiology and Orthopedics, the stable results (as evidenced by the narrow standard deviations and significant p-values) highlight the model's ability to generalize to multi-center data and heterogeneous data distributions. While improvements on some benchmarks appear modest, the combination of high accuracy, expert-level knowledge integration, and robust cross-modal reasoning provides a tangible benefit for real-time clinical decision support.

\section{Conclusion}
In this paper, we propose a knowledge graph enhanced cross-Mamba interaction (KG-CMI) framework, which demonstrates significant advancements in handling both closed-ended and open-ended medical visual question answering tasks. KG-CMI integrates a fine-grained visual-text feature alignment module, a knowledge graph embedding module, a cross-modal interaction representation module, and a free-form answer enhanced multi-task learning module. The framework effectively captures the complex correlations between medical images, text questions, and medical knowledge, enabling the model to reason about medical concepts and relationships more effectively.
Through experiments on multiple benchmark datasets, including SLAKE, VQA-RAD, and OVQA, our model has demonstrated excellent performance. Furthermore, interpretability analysis underscores the model's ability to focus on relevant image regions when answering questions, enhancing its practical applicability in real-world medical scenarios.

\section*{References}
%
%
\bibliographystyle{IEEEtran}
\bibliography{ref}
\end{document}